\documentclass{article} % For LaTeX2e
\usepackage{iclr2025_conference,times}

\usepackage{amsmath,amsfonts,bm}

\def\eqref#1{equation~\ref{#1}}
\def\1{\bm{1}}

\DeclareMathAlphabet{\mathsfit}{\encodingdefault}{\sfdefault}{m}{sl}
\SetMathAlphabet{\mathsfit}{bold}{\encodingdefault}{\sfdefault}{bx}{n}

\DeclareMathOperator*{\argmax}{arg\,max}

\usepackage{hyperref}
\usepackage{url}
\usepackage{tcolorbox}
\usepackage{booktabs}
\usepackage{multirow}
\tcbuselibrary{skins}

\newcommand{\algname}{EmbedLLM}

\title{\algname{}: Learning Compact Representations of Large Language Models}

\iclrfinalcopy % Uncomment for camera-ready version, but NOT for submission.
\author{Richard Zhuang\thanks{Equal contribution.}\quad Tianhao Wu\footnotemark[1] \quad Zhaojin Wen \quad Andrew Li\\ \textbf{Jiantao Jiao} \quad \textbf{Kannan Ramchandran}
\\
\\
\textsuperscript{}University of California, Berkeley \quad
}

\usepackage{tablefootnote}
\usepackage{enumitem}
\usepackage{bbm}
\usepackage{amsmath}

\newtcolorbox{prompt}[1]{
    enhanced,
    drop shadow=black!5!white,
    left=4mm,
    right=4mm,
    top=2mm,
    bottom=2mm,
    boxsep=0mm,
    rounded corners,
    title=#1,
    fontupper=\footnotesize\linespread{0.9}\fontfamily{lmr}\selectfont,
    }

\begin{document}

\maketitle
\begin{abstract}
With hundreds of thousands of language models available on Huggingface today, efficiently evaluating and utilizing these models across various downstream tasks has become increasingly critical. Many existing methods repeatedly learn task-specific representations of Large Language Models (LLMs), which leads to inefficiencies in both time and computational resources. To address this, we propose \algname{}, a framework designed to learn compact vector representations of LLMs that facilitate downstream applications involving many models, such as model routing. 
%Through a reconstruction-based approach, \algname{} embeds models into a lower-dimensional space, preserving essential characteristics required for tasks such as correctness forecasting, routing, and accuracy benchmarking. 
We introduce an encoder-decoder approach for learning such embeddings, along with a systematic framework to evaluate their effectiveness.
%we also establish an evaluation setup that leverages the learned embedding to predict model behavior on unseen tasks, reducing the need for exhaustive per-task model retraining. 
Empirical results show that \algname{} outperforms prior methods in model routing both in accuracy and latency. Additionally, we demonstrate that our method can forecast a model's performance on multiple benchmarks, without incurring additional inference cost. Extensive probing experiments validate that the learned embeddings capture key model characteristics, \textit{e.g.} whether the model is specialized for coding tasks, even without being explicitly trained on them.
%Our approach offers a scalable way to navigate the growing landscape of LLMs, \textit{i.e. through an vector embedding}. 
We open source our dataset, code and embedder to facilitate further research and application: \url{https://github.com/richardzhuang0412/EmbedLLM}. 
\end{abstract}

\section{Introduction}

Recent breakthroughs in Large Language Models (LLMs) \citep{vaswani2023attentionneed} have led to the creation of a vast array of models, each tailored for different use cases. These models, ranging from small, specialized models to large, general-purpose systems \citep{hao2022languagemodelsgeneralpurposeinterfaces}, differ significantly in their architecture, size, training data, and performance characteristics. For example, while some models excel as conversational agents, others may be more suitable for code generation or logical reasoning tasks.  However, with this explosion of diverse LLMs comes a major challenge: 
\begin{center}
\textit{How to efficiently manage, compare, and utilize the growing number of LLMs?}
\end{center}

\begin{figure}[h]
    \centering
    \includegraphics[width=1.0\linewidth]{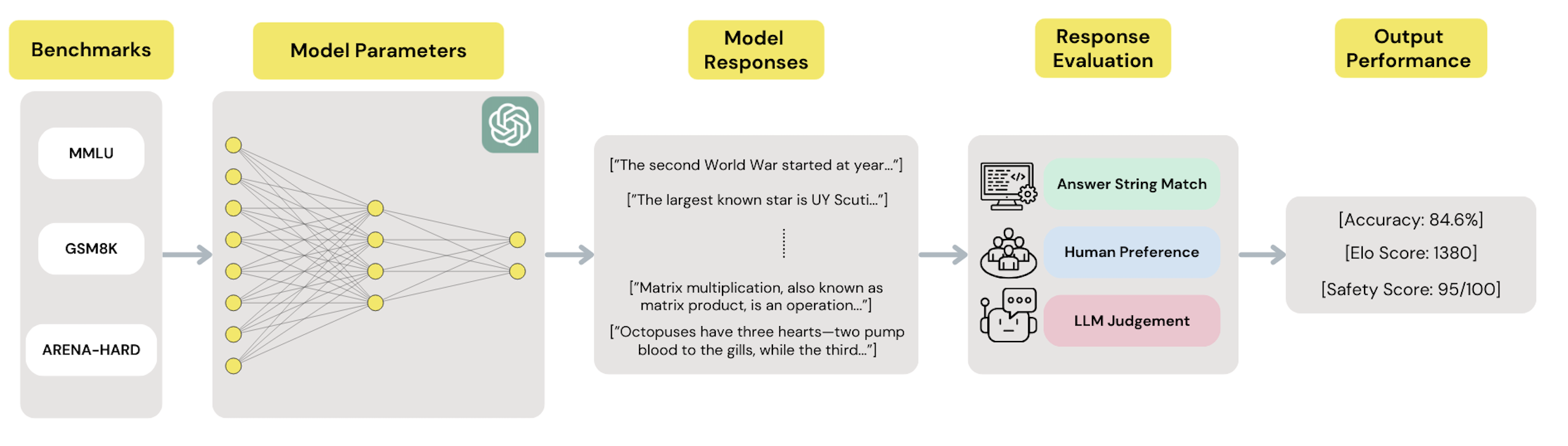}
    \caption{An illustration of the traditional workflow of LLM benchmarking.}
    \label{fig:benchmarking_diagram}
\end{figure}

\begin{figure}[h]
    \centering
    \includegraphics[width=0.95\linewidth]{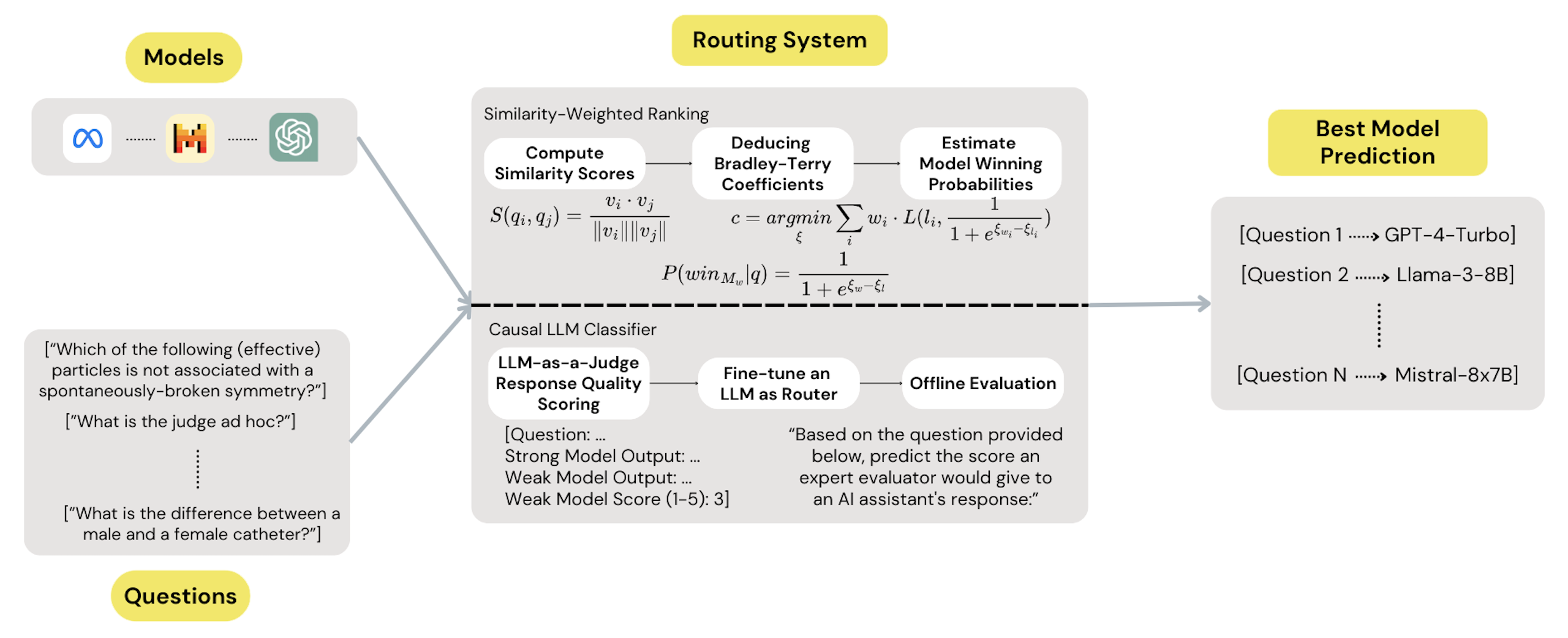}
    \caption{An illustration of the traditional workflow of model routing, using exemplar routing methodologies from \cite{ong2024routellm}}
    \label{fig:routing_diagram}
\end{figure}

Traditionally, benchmarking has served as the primary method for comparing LLMs, where each model is evaluated on a fixed set of test cases, and a score is generated to represent its performance. Meanwhile, model routing systems are developed to efficiently select models given queries of different types. An example workflow of these tasks can be seen in \autoref{fig:benchmarking_diagram} and \autoref{fig:routing_diagram}. While these approaches are often robust indicators of a model’s strengths and weaknesses, the construction of their workflows induces repeatedly learning representations of various LLMs to suit each individual downstream tasks and is therefore time-consuming and compute-demanding.

In response to these challenges, we introduce \textbf{\algname{}}, a compute-friendly framework designed to learn compact vector representations of large language models that facilitates different tasks. \algname{} map models into a latent vector space that captures important model characteristics. More importantly, \algname{} produces a unified representation that can be simultaneously applied to various downstream tasks such as correctness forecasting (\autoref{sec:correctness_forecast}), model routing (\autoref{sec:routing}), and benchmark accuracy evaluation (\autoref{sec:bench_pred}). The core idea is to enforce this representation learning through a reconstruction-based system that tries to predict the model's answer (correctness) from the learned embeddings, ensuring that each model’s embedding retains the most salient features relevant to enhance performance across multiple scenarios.

% One of the major contributions of our work is the introduction of this reconstruction system to reduce the dimensionality of LLM representations while maintaining their essential features. Along with the embedding framework, we also provide a comprehensive evaluation setup that leverages these embedding to predict model behavior on unseen prompts without requiring full re-evaluation on every task. This allows for efficient task routing and correctness forecastings based on the learned embedding (it would be good to cite related work that addresses evaluation setups for model behavior prediction, like papers discussing evaluation harnesses or model probing techniques).

% Furthermore, we conduct extensive probing experiments to validate the effectiveness of the learned embedding. These experiments demonstrate that \algname{} not only captures model correctness but also reflects deeper, task-specific characteristics, ensuring that the representations are both information-rich and versatile. By systematically exploring the embedding, we show that our framework successfully differentiates between models based on their ability to handle complex reasoning, mathematical tasks, and domain-specific queries (it would be good to cite probing experiments for evaluating model embedding, particularly those used in language models).
\begin{figure}[h]
    \centering
    \includegraphics[width=0.95\linewidth]{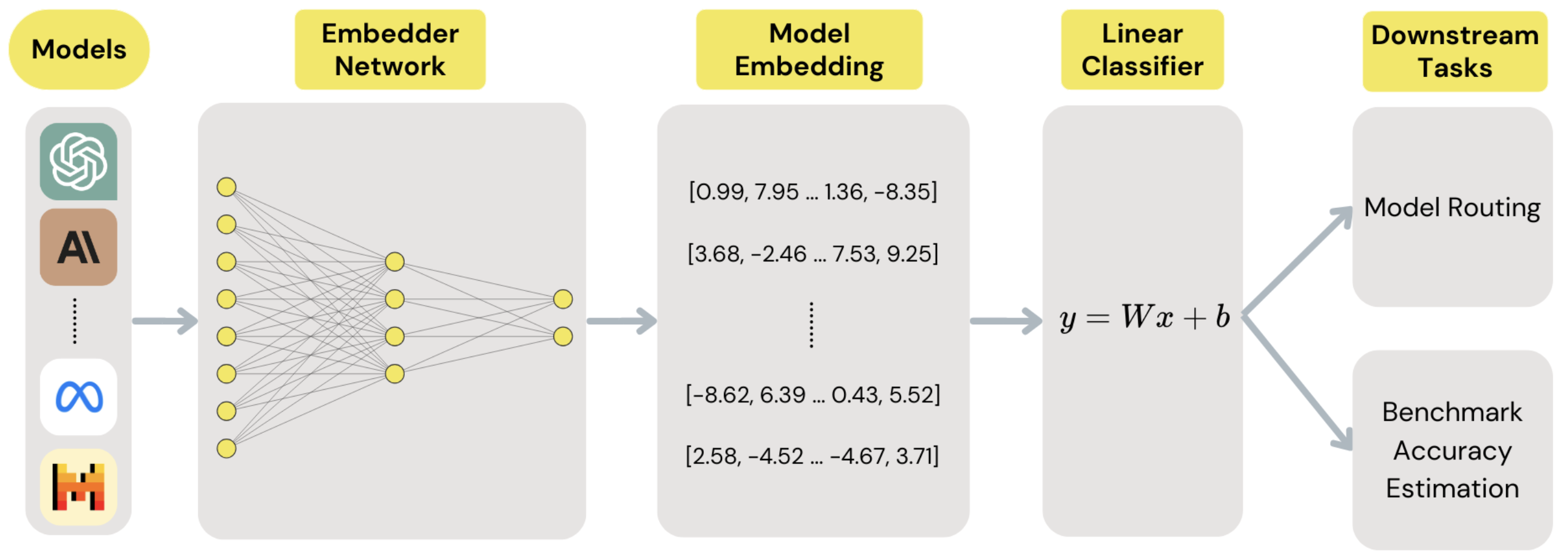}
    \caption{An illustration of the \algname{} Pipeline. An embedder network is pretrained to convert models into vector embeddings. Downstream applications like model routing are adapted by training an additional linear layer on top of these embeddings.}
    \label{fig:EmbedLLM_diagram}
\end{figure}

In summary, our contributions are as follows:
\begin{enumerate}
    \item We propose a novel framework based on a reconstruction system for learning compact, low-dimensional representations of LLMs. To the best of our knowledge, this is the first work that explicitly introduces the concept of turning LLMs into embeddings.
    \item We introduce an evaluation setup that employs the model embeddings to \emph{simultaneously} predict model answer correctness on unseen questions, perform model routing, and predict benchmark accuracy  with the addition of only a linear classifier.
    %only additional linear classifier involved. 
    Our system demonstrates the potential for significantly reducing the need for task-specific re-training.
    \item We perform probing experiments to validate that the learned embeddings capture meaningful information. We discover that models with similar characteristics remain close in the embedding space, and that the effect of incorporating each benchmark is reflected through the change in model embeddings. %These results further confirms the embedding's utility for various downstream tasks.
\end{enumerate}

\algname{} offers a scalable and unified approach to model evaluation and selection. By producing embeddings that encapsulate important features across tasks, our framework provides a versatile method to navigate the increasingly complex landscape of large language models.

\section{Related Work}

\textbf{Representation Learning}  There have been numerous attempts to learn representations of various types of information. For natural languages, \cite{mikolov2013efficientestimationwordrepresentations} and \cite{pennington-etal-2014-glove} revolutionized the way models capture word semantics. In the field of computer vision, self-supervised techniques \citep{noroozi2017unsupervisedlearningvisualrepresentations} \citep{vondrick2018trackingemergescolorizingvideos} are designed to learn low-dimensional representations that bolster downstream classification or segmentation performances. Inspired by these work and realizing an increasing demand of various LLMs being trained, we propose a creative framework to learn embeddings of LLMs.

\textbf{LLM Benchmarking} Benchmarking has been a standard way to compare LLMs, where a collection of questions/prompts is input to the LLMs and the quality of the corresponding responses is evaluated. There are datasets curated for general knowledge and language understanding \citep{wang2019gluemultitaskbenchmarkanalysis}, as well as domain-specific ones evaluating reasoning abilities \citep{frohberg2022crassnoveldataset}, narrative control \citep{hartvigsen2022toxigenlargescalemachinegenerateddataset}, function-calling \citep{berkeley-function-calling-leaderboard,srinivasan2023nexusraven} and instruction-following \citep{li2024crowdsourced,dubois2024length}. However, assessing quality of benchmarks remains an under studied question, and systematically evaluating all of them incurs an enormous amount of inference cost.

\textbf{LLM Routing} Our work focuses on predictive routing, which is a technique aimed at proposing the most suitable model given a task, without actually passing the query through each one of them. As summarized in \cite{hu2024routerbenchbenchmarkmultillmrouting}, most routers adopt either a supervised technique \citep{ong2024routellm,shnitzer2023largelanguagemodelrouting} or a reward-based method \citep{lu2023routingexpertefficientrewardguided}. Trained by seeing responses from different models to the same prompt, these systems are intrinsically building an understanding of key model characteristics. Our work establishes yet another new interesting and efficient research direction as we find model 
embeddings strengthen routing performances.

\section{Formulation}

\subsection{Problem Setup}
Let $\mathcal{M} = \{M_1, M_2 \cdots M_n\}$ be a set of different LLMs, $\mathcal{P}$ denote the set of all possible prompts, and $\mathcal{A}$ denote the corresponding set of possible answers. We can simply identify any LLM $M$ with an inference function mapping from a prompt space to an answer space $f_M: \mathcal{P} \rightarrow \mathcal{A}$, which outputs an answer $a \in \mathcal{A}$ given a prompt $p \in \mathcal{P}$. 

Among downstream tasks, representations of different LLMs needed to be constructed in various ways. A naive example is benchmarking: Where the crucial part is to select a test prompt set $\mathbb{P}_{Bench} = \{p_1, p_2 \cdots p_m \}$ as well as an scoring function $g_{eval}: \mathcal{P}\times\mathcal{A} \rightarrow \mathbb{R}$, mapping model responses to a scalar score. Take MMLU as an example, each model is queried by a set of multiple-choice questions with four choices. The model's response is recorded by comparing the output probabilities of the answer choices ``A", ``B", ``C", and ``D". Then the responses are simply scored by matching with the ground truth answer key. Within this process, every model $M$'s behavior is essentially summarized by their output on a set of test questions$\{(p_1, a_{(M,1)}), (p_2, a_{(M,2)}) \cdots (p_m, a_{(M,m)}) \}$. 

Another example is model routing: given a pool of $n$ LLMs, a router function is defined in \cite{ong2024routellm} as a $n$-way classifier that assigns models to different queries to maximize response quality while minimize inference cost. Similarly, training such router also often involves in transforming each model into a relatively low-dimensional representation by utilizing sample question-answer pairs. 

Identifying a common need for model characterization, we raise the question: what if we could accomplish the above tasks by \textbf{directly working with a unified, compact representation of LLMs}? In this work, we provide a framework to learn such a representation. We define an embedding function $\phi : \mathcal{M} \rightarrow \mathbb{R}^d$, parametrized by a neural network or otherwise, that maps each model $M$ to a compact vector representation in a latent embedding space. We aim to learn model embeddings that contain important features of LLMs that are useful to both quantifying differences between various models and aiding across downstream tasks.

\subsection{Evaluation Metrics}
\label{sec:eval_metrics}
Evaluating the quality of model embeddings is crucial to ensure they effectively capture the underlying structure and semantics of the data which we care about for potential downstream tasks. The core idea is to use the embeddings to predict model behavior on unseen tasks by training an inference function $\psi : \mathbb{R}^d \times \mathcal{P} \rightarrow \mathbb{Q}$, that leverages a model embedding and new test prompts to predict a model's performances as quantified by a desired evaluation metric in the space $\mathbb{Q}$. For instance, if our task is to predict whether a model can correctly answer an unseen question, the inference function would be a network that takes in model embeddings and the unseen question (usually present in the form of its embedding as well) and output a binary label as the prediction of model correctness. Note that model responses can be evaluated through many means such as determining correctness \citep{hendryckstest2021MMLU}, judging by human \citep{chiang2024chatbotarenaopenplatform} or stronger LLMs \citep{zheng2023judging}, or measuring any other task-specific metric. Hence, each task constitute to its own unit of measure and correspondingly determines an evaluation metric. In this work, we focus on evaluating on the following downstream tasks:

\begin{itemize}
    \item \textbf{Correctness Forecasting:} We query the models on benchmarks equipped with ground truth answers, and produce binary correctness label for every model-question pair. For this task, a natural metric to use would be test prediction accuracy.
    \item \textbf{Model Routing:} With the help of a embedding vector of each model, we're able to develop a simple linear router that directly determine the model to be routed to using the probability of the answer correctness along with some learnable threshold. We measure the routing performance by reporting the average response quality\footnote{Note that this metric can be insufficient under certain settings \citep{ong2024routellm}, but it fits well to our goal of assessing whether our model embeddings effectively captures the strengths and weaknesses under the context of correctness forecasting.}, namely the proportion of times a router successfully route to an LLM that correctly answers the given query.
    \item \textbf{Benchmark Accuracy Prediction:} We treat the embeddings as primary features to train a linear model in predicting accuracies on unseen benchmarks (measured from 0\% to 100\% as a percentage). The metric used for this task would be the classical mean-squared-error (MSE) for linear regression.
\end{itemize}

In addition, we could directly compare between embeddings in their raw vector form following \cite{mikolov-etal-2013-linguistic}. For instance, models with similar traits should have embeddings that are closer in L2 distance. 

\section{\algname{}}

\subsection{Methodology}

In order to learn such embeddings, we draw inspirations from image reconstruction algorithms \citep{He_2022_CVPR_MAE,ronneberger2015unetconvolutionalnetworksbiomedical}: we want to learn a ``reconstruction'' system, where the choice of the reconstruction target is arbitrary and can be task-dependent. More concretely, let $\mathbb{Q}$ be a space of model performance metric, and $\mathcal{M}$ be a set of possible LLMs, $m$ be the number of models, $n$ be the number of questions/prompts. We want to learn a reconstruction network $\mathcal{R}: \mathbb{Q}^{m \times n} \rightarrow \mathbb{Q}^{m \times n}$ that optimizes to reconstruct a matrix $X \in \mathbb{Q}^{m \times n}$, where $X_{ij} \in \mathbb{Q}$ denotes the model performance metric for model i on question j. The encoder-decoder architecture ensures the imposure of such constraint and enforce the model embeddings to efficiently capture the key characteristics of each model.

In this work, we decide to use the task of predicting model answer correctness as our auxiliary target, \textit{i.e.}, $\mathbb{Q} = \{0,1\}$. Notice that our training objective is only a decoy - any downstream task that requires understanding of model characteristics would qualify and our ultimate goal is to enforce the reconstruction network to learn a compact yet information-rich representation of the models in this process.

\subsection{Dataset}
\label{sec:correctness_data}

We now describe the data collection process of correctness results of various LLMs' responses to questions from mainstream benchmarks. 

We selected 112 open-sourced models\footnote{See appendix for full list.} of various sizes, with both general purpose LLMs \citep{zhu2024starling} and specialized LLMs included to ensure comprehensive coverage. Then we aggregated responses of every model to 36,054 questions from the test sets of MMLU \citep{hendryckstest2021MMLU}, TruthfulQA \citep{lin2022truthfulqameasuringmodelsmimic} , SocialQA \citep{sap2019socialiqacommonsensereasoningsocial}, PIQA\citep{bisk2019piqareasoningphysicalcommonsense}, MedMCQA\citep{medmcqa}, MathQA\citep{amini-etal-2019-mathqa}, LogiQA\citep{liu2020logiqachallengedatasetmachine}, GSM8K\citep{cobbe2021gsm8k}, GPQA\citep{rein2023gpqagraduatelevelgoogleproofqa}, and ASDiv\citep{asdiv}. The responses to these questions were acquired and evaluated through using the ``lm-evaluation-harness'' package \citep{eval-harness} to give a binary correctness label for each model-question pair. We performed a random 80\%-10\%-10\% train-validation-test split on the questions and used the sentence transformer ``all-mpnet-base-v2'' \citep{reimers-2019-sentence-bert} to convert the questions into an initial embedding state of dimension $dim_q = 768$. Consequently, our question embedding tensor $X$ has the shape (36054, 768) where the $i$-th row $X_i \in \mathbb{R}^{dim_q}$ is the embedding for model $i$ , and our label tensor is essentially a correctness matrix $Y$ with shape (112, 36054), where the \(ij\)-th entry $Y_{ij}$ represents the binary correctness of model \(i\) answering question \(j\).

\subsection{Algorithm}

We adopt the Encoder Decoder architecture:

\textbf{Encoder:} The encoder consists of a model embedding network and a question embedding network. Let $dim_{embed}$ be the desired dimension of the model embedding. The model embedding network $\phi_{m}: \mathcal{M}$\footnote{In the actual implementation, we give each model an index as an identification in $\mathcal{M}$.} $\rightarrow \mathbb{R}^{dim_{embed}}$ maps each model into a latent representation. Similarly, the question embedding network $\phi_{q}: \mathcal{P} \rightarrow \mathbb{R}^{dim_{embed}}$ maps each questions into a latent representation. In our setting, the question embedding network is a two-step transformation $\phi_{q} = g_{st} \circ h_{proj}$ where $g_{st}: \mathbb{P} \rightarrow \mathbb{R}^{dim_q}$ denotes the pre-processing (performed in \autoref{sec:correctness_data}) that turns each question from text to an initial embedding space by using sentence transformer, and $h_{proj}: \mathbb{R}^{dim_q} \rightarrow \mathbb{R}^{dim_{embed}}$ is a projection layer from this original embedding space to the same space as the model embeddings.

\textbf{Decoder:} The decoder is a classifier network $\psi: \mathbb{R}^{dim_{embed}} \times \mathbb{R}^{dim_{embed}} \rightarrow \{0,1\}$ that takes in both the encoded embeddings of the model and the question, and output a binary label which is the prediction of whether the model answer the question correctly. For this work, our decoder is represented as $\psi(v_{m}, v_{q}^{'}) = \varphi(v_{m} \odot v_{q}^{'})$ where $\varphi: \mathbb{R}^{dim_{embed}} \rightarrow \mathbb{R}^{2}$ is a linear classifier and $u \odot v$ represents the Hadamard (element-wise) product between two vectors. For each model-question pair, this decoder network outputs two logits $p_{{(m,q)}_{0}}$ and $p_{{(m,q)}_{1}}$, and the ``correctness score" $s_{m,q} = \sigma  (p_{{(m,q)}_{1}} - p_{{(m,q)}_{0}})$ represent the predicted probability of the model $m$ correctly answering question $q$, where $\sigma(x)$ is the sigmoid function. 

Suppose $y$ is the correctness label of model $m$ answering question $q$, we calculate the following BCE loss function during training,
\begin{equation}
L(m,q,y) = - \left( y \cdot \log(s_{m,q}) + (1 - y) \cdot \log(1 - s_{m,q}) \right)
\end{equation}

 %During inference, we take the argmax of the logits as the output label. We use the correctness score to route between model as explained in more detail in Section \ref{sec:routing}. 
 In essence, this algorithm analogizes to a \textbf{Matrix Factorization} algorithm, where we learn a $n \times m$ model embedding matrix and a $m \times p$ question embedding matrix such that their product reconstructs the original $n \times p$ correctness matrix. Hence, we refer to this reconstruction system as ``Matrix Factorization'', or ``MF'' in short, in the following sections. 

\section{Experiment Results}

As described in \autoref{sec:eval_metrics}, we conducted experiment in correctness forecasting, model routing, and benchmark accuracy prediction to evaluate the quality of the learned embeddings. 

\subsection{Correctness Forecasting}
\label{sec:correctness_forecast}

For correction prediction, we compare the effect of our matrix factorization model to a KNN-classifier \citep{fix1985discriminatory}. In the context of our formulation, although without an explicit embeddings for models, KNN-classifier can be seen as using the integration of all question-correctness tuples from a model as its ``embedding'', and making inference from this aggregation. For brevity, we refer to this approach as KNN in the subsequent text. As mentioned in \autoref{sec:eval_metrics}, we use correctness forecasting accuracy on the test set as the evaluation metric.

We evaluate the performance of KNN and Matrix Factorization across various sizes of training set. We produce the smaller training sets from randomly subsetting from the full training set. For each training set, we conduct hyperparameter tuning (number of neighbors for KNN, model embedding dimension for MF) on a fixed validation set and evaluate prediction accuracy using a fixed test set. The result in \autoref{table:scale_experiment} indicates a better scalability of our method.

% \begin{table}[h!]
% \centering
% \begin{tabular}{ c|c|c|c|c|c|c|c }
%  \hline
%  Algorithm & 1K & 5K & 10K & 15K & 20K & 25K & Full (29K)\\
%  \hline
%  KNN & 0.6372 & 0.7078 & 0.7107 & 0.7128 & 0.7143 & 0.7146 & 0.7152\\
%  \hline
%  Matrix Factorizaton & 0.6443 & 0.7331 & 0.7362 & 0.7378 & 0.7390 & 0.7394 & $\mathbf{0.7409}$ \\
%  \hline
% \end{tabular}
% \caption{Performance of Predicting Model Correctness on Unseen Questions where the columns indicates the number of questions in the training set. MF constantly outperforms KNN on training sets of all scale. }
% \label{table:scale_experiment}
% \end{table}

\begin{table}[h!]
\centering
\begin{tabular}{ c|c|c|c|c|c|c|c }
 \hline
 \multirow{2}{*}{\textbf{Algorithm}} & \multicolumn{7}{c}{\textbf{Dataset Size}} \\
 \cline{2-8}
 & \textbf{1K} & \textbf{5K} & \textbf{10K} & \textbf{15K} & \textbf{20K} & \textbf{25K} & \textbf{Full (29K)} \\
 \hline
 KNN & 0.6372 & 0.7078 & 0.7107 & 0.7128 & 0.7143 & 0.7146 & 0.7152\\
 \hline
 Matrix Factorization & 0.6443 & 0.7331 & 0.7362 & 0.7378 & 0.7390 & 0.7394 & $\mathbf{0.7409}$ \\
 \hline
\end{tabular}
\caption{Performance of Predicting Model Correctness on Unseen Questions where the columns indicate the number of questions in the training set. MF constantly outperforms KNN on training sets of all scale. We see a steady increase in the performance as the dataset size grows, indicating further scalability of our method.}
\label{table:scale_experiment}
\end{table}

% \begin{figure}[h]
%     \centering
%     \includegraphics[width=0.75\linewidth]{figs/correctness_prediction.png}
%     \caption{Performance of Predicting Model Correctness on Unseen Questions. MF constantly outperforms KNN on training sets of all scale.}
%     \label{fig:correctness_prediction}
% \end{figure}

\subsection{Model Routing}
\label{sec:routing}
Using the same correctness data, we can evaluate the quality of a matrix-factorization-based router. For this task, we evaluate the router's accuracy by measuring the proportion of times it successfully route to a model that could correctly answer the given query. For a test question $q_k$, we pass it through the router network along with all $n$ possible model embeddings, producing $n$ correctness score $s_{M_1,q_k}, s_{M_2,q_k}, \cdots, s_{M_n,q_k}$, and check if the model with the highest correctness score correctly answers the question. Aggregating through all questions, we report the router accuracy as $acc_{router} = \frac{1}{m} \sum_{k = 1}^{m} \mathbbm{1}\{X_{i^{*},k} = 1\} $ where $i^{*} = {\argmax} \{s_{M_i,q_k} | i = 1 \cdots n\}$ is the routed model and $X$ is the correctness matrix described in \autoref{sec:correctness_data}.

We compare the performance of the Matrix Factorization router with two baselines. 
The first one is the single-best model router which always selects the model with the highest accuracy and thus gives a constant accuracy. The second one is a random router that select each model the same number of times as the Matrix Factorization router, but instead randomly assign models to questions. For this router, we can calculate its expected accuracy given the proportions of times each model is selected by the Matrix Factorization router. For instance, if our router selects $M_1$ 70\% of the time, $M_2$ 20\% of the time, and $M_3$ 10\% of the time, the expected accuracy of the random router will be calculated as a weighted accuracy $acc_{weighted} = 0.7 * acc_{M_1} + 0.2 * acc_{M_2} + 0.1 * acc_{M_3}$. Note that this weighted accuracy will always be smaller than the single-best model accuracy - we propose this metric as an evaluation of how well our router direct models to the questions they are good at given a fixed ``budget'' of model calls. We report both the overall accuracy across the whole test set and accuracy per source benchmark. 

\begin{figure}[h]
    \centering
    \includegraphics[width=0.8\linewidth]{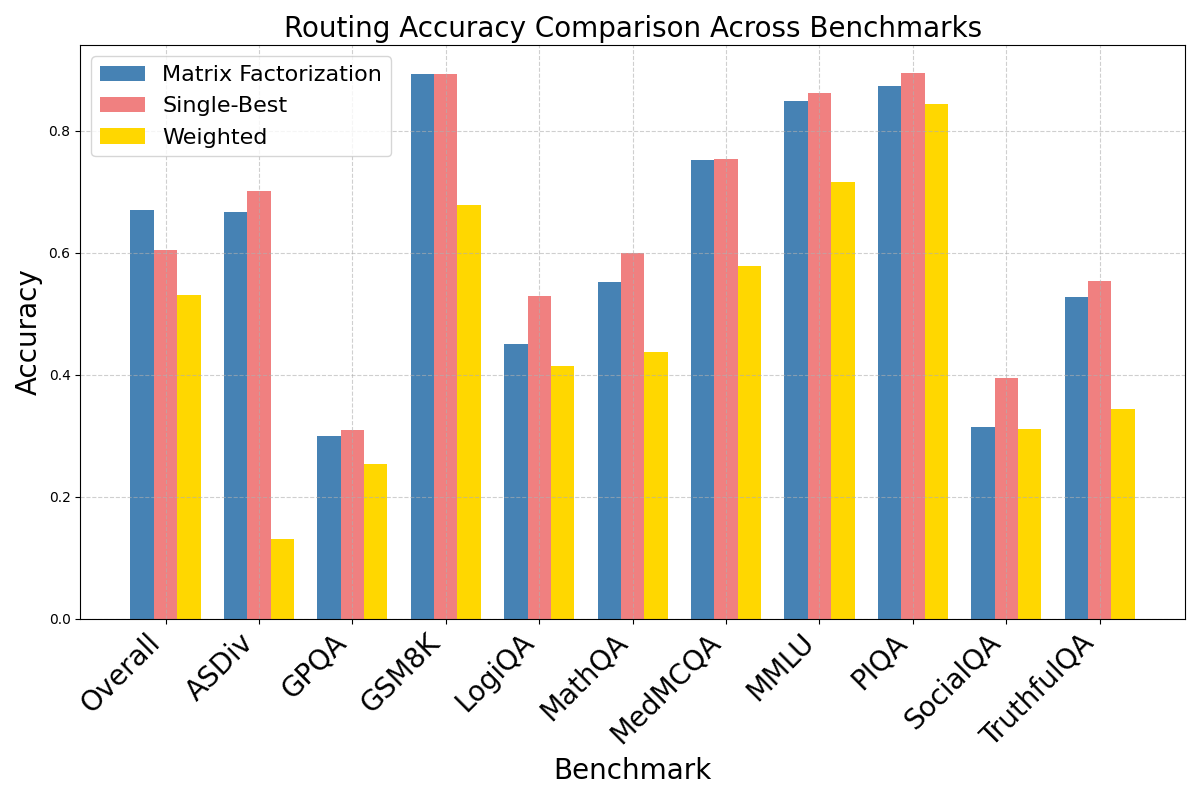}
    \caption{Performance accuracy of MF router compared to baselines. MF router performs better across the whole test set and achieves accuracies close to the single-best model on every benchmark.}
    \label{fig:routing_accuracy}
\end{figure}

As seen in \autoref{fig:routing_accuracy}, MF router performs better than both the single-best router and the random router overall. As the best performing model is re-determined on every benchmark, the single-best model router performs better for each benchmark than in overall case. Here, MF router achieves near single-best model router accuracy while still managing to utilize the respective strengths of different models which is shown by the significant difference between accuracies of MF router and weighted router.

Another advantage of MF router we find is its high routing speed. On one NVIDIA A100 80GB GPU, it takes in average 3.80 seconds for MF router to route 3,000 questions on 50 repeated trials which is basically free compare to the downstream model inference time. Specifically, compared to the causal LLM router in \cite{ong2024routellm} that processes less than 50 requests per second, our router delivers more than 750 model selections, which is \textbf{15x faster} while selecting from a much larger model pool (112 models against 2 models for causal LLM router). In addition, it only takes less than 1GB of GPU memory when training a MF router using our dataset, which is \textbf{60x cheaper} than fine-tuning Llama-3-8B as a router in terms of memory usage. This illustrates the superiority of MF router in both performance, latency, and training cost. 
% \begin{table}[h!]
% \centering
% \begin{tabular}{ c|c|c|c }
%  % \hline
%  Benchmark & Matrix Factorization & Single-Best & Weighted\\
%  \hline
%  Overall & 0.6697 & 0.6050 & 0.5304 \\
%  \hline
%  ASDiv & 0.6667 & 0.7020 & 0.1298 \\
%  \hline
%  GPQA & 0.3000 & 0.3100 & 0.2528 \\
%  \hline
%  GSM8K & 0.8929 & 0.8929 & 0.6783 \\
%  \hline
%  LogiQA & 0.4510 & 0.5294 & 0.4142 \\
%  \hline
%  MathQA & 0.5527 & 0.5992 & 0.4378 \\
%  \hline
%  MedMCQA & 0.7514 & 0.7542 & 0.5783 \\
%  \hline
%  MMLU & 0.8489 & 0.8625 & 0.7168 \\
%  \hline
%  PIQA & 0.8731 & 0.8955 & 0.8441 \\
%  \hline
%  SocialQA & 0.3148 & 0.3951 & 0.3115 \\
%  \hline
%  TruthfulQA & 0.5270 & 0.5541 & 0.3439 \\
%  \hline
% \end{tabular}
% \caption{Performance of Matrix Factorization based router against two baselines.}
% \label{table:routing_experiment}
% \end{table}

\subsection{Benchmark Accuracy Prediction}
\label{sec:bench_pred}

To predict model's average accuracy on a benchmark $B$, we trained MF using leave-one-out correctness data, which includes correctness results of all models on all questions except the ones in $B$. Then we take the model embeddings directly as features to train a linear regression of the form:
\begin{equation*}
a \mathbf{E} = \mathbf{y}
\end{equation*}
where \(E\) is a model embedding matrix with the \(i\)-th row representing the model embedding for the \(i\)-th model, and the $j$-th entry in the vector \( \mathbf{y} \) corresponds to the \(j\)-th model’s average correctness accuracy on the test benchmark, which is a number from 0 to 1.

\begin{figure}[h]
    \centering
    \begin{minipage}[t]{0.45\textwidth}
        \centering
        \begin{tabular}{|l|r|}
            \hline
            \textbf{Benchmark} & \textbf{Significance} \\
            \hline
            MathQA         & 100 \\
            LogiQA         & 100 \\
            MedMCQA        & 100 \\
            PIQA           & 98  \\
            TruthfulQA     & 96  \\
            MMLU           & 94  \\
            GSM8K          & 93  \\
            GPQA           & 10  \\
            ASDiv          & 6   \\
            SocialQA       & 3   \\
            \hline
        \end{tabular}
        % \caption{Sorted Kendall's Tau test result of the benchmarks. The ``Significance'' column represents the number of times with significant correlation detected (at a 5\% significance level) out of 100 random model splits.}
        \label{tab:kt_test}
    \end{minipage}
    % \hfill
    \begin{minipage}[t]{0.50\textwidth}
        \vspace{-27.5mm}
        \centering
        \includegraphics[width=\linewidth]{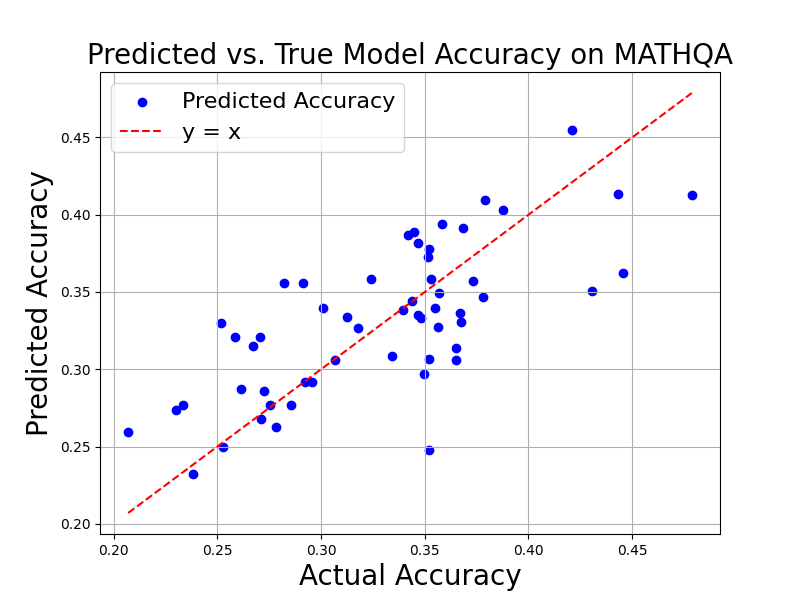}
        % \caption{An example comparing actual model accuracies on MathQA against model accuracies on MathQA predicted from the embedding trained without MathQA data.}
        \label{fig:regression_example}
    \end{minipage}
            \caption{\textbf{Left:} Sorted Kendall's Tau test result of accuracy prediction on the benchmarks. The ``Significance'' column represents the number of times with significant correlation detected (at a 5\% significance level) out of 100 random model splits. \textbf{Right:} An example comparing actual model accuracies on MathQA against model accuracies on MathQA predicted from the embeddings trained without MathQA data.}
\end{figure}

For each test benchmark, we conducted 100 random train-test splits on the 112 models contained in our dataset, trained a linear regression on the training set, and evaluated the correlation between model embeddings and test benchmark performances on the test set through applying Kendall's Tau test\footnote{The Kendall's Tau test is a measure of correspondence between two rankings. We use this test to see if model ability can be correctly ordered simply using a linear system with the embeddings as the only feature. }. From \autoref{tab:kt_test}, statistical significance is found in 7 out of the 10 benchmarks, indicating that model embedding contains information to distinguish between model performances on most benchmarks.

% \begin{table}[h]
%     \centering
%     \begin{tabular}{|l|r|}
%         \hline
%         \textbf{Benchmark} & \textbf{Significance} \\
%         \hline
%         MathQA         & 100 \\
%         LogiQA        & 
%         100 \\
%         MedMCQA    & 
%         100 \\
%         PIQA       & 
%         98 \\
%         TruthfulQA         & 96 \\
%         MMLU        & 
%         94 \\
%         GSM8K          & 
%         93 \\
%         GPQA        & 10 \\
%         ASDiv          & 6 \\
%         SocialQA        & 3 \\
%         \hline
%     \end{tabular}
%     \caption{Sorted Kendall's Tau test result of the benchmarks. The ``Significance'' column represents the number of times with significant correlation detected (at a 5\% significance level) out of 100 random model split.}
%     \label{tab:kt_test}
% \end{table}

% \begin{figure}[h]
%     \centering
%     \includegraphics[width=0.5\linewidth]{figs/regression_example_mathqa.png}
%     \caption{An example comparing actual model accuracies on MathQA against model accuracies on MathQA predicted from the embedding trained without MathQA data.}
%     \label{fig:regression_example}
% \end{figure}

Notice that this prediction systems works even for large benchmarks like MMLU. Using correctness result from the rest of the 18,000 questions, our method predicts MMLU accuracy to a statistically significant extent. This saves an inference cost of 14,000 questions, equivalent to around 2 million input tokens per model, let alone cost for model output tokens and subsequent response evaluation. As number of models, model sizes and number of benchmarks are still rising rapidly, enabling benchmark accuracy prediction through model embeddings is vital to save both time and compute from repeatedly inferencing models on every new benchmark.

\section{What information is in the Model embeddings}

In this section we describe the probing experiments designed to understand what information is captured in the embedding.

\subsection{Sanity Check Using Similarity}

We expect the model embeddings to satisfy some basic properties: If two models $M, M'$ generate the same answers for every prompt, then their embeddings are the same. Similarly, models with similar characteristics, trained using similar data, or adopted similar training pipelines should have similar embeddings, and vice versa. For instance, the model embedding of DeepSeekMath-7B \citep{shao2024deepseekmathpushinglimitsmathematical} should be more similar to the embedding of other math models like MetaMath-Llemma-7B \citep{yu2023metamath} than to the embedding of Medicine-LLM-13B \citep{cheng2024adapting} which is adapted for biomedical applications. This property is easily fulfilled by Matrix Factorization algorithm as any two identical/similar models of such would produce identical/similar correctness result against most questions. 

\begin{figure}[h]
    \centering
    \includegraphics[width=0.7\linewidth]{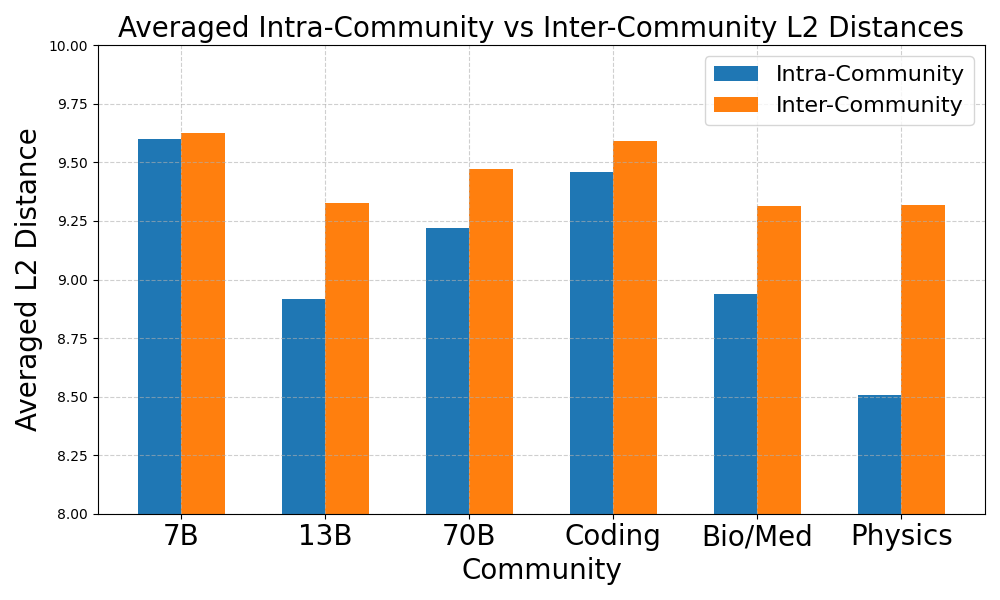}
    \caption{The averaged intra-community L2 distance of the model embeddings is closer for all 6 communities selected, suggesting that basic model traits are captured in the latent embedding space.}
    \label{fig:community_distance}
\end{figure}

As a further sanity check, we assign binary labels to the 112 models we have evaluated according to the following 6 keywords: [7B, 13B, 70B, Coding, Bio/Med, Physics], forming 6 characteristic communities. For each community, we compare between the average intra-community and inter-community L2 distance of the embeddings. As shown in \autoref{fig:community_distance}, for all above 6 communities, the averaged intra-community L2 distances are smaller than the inter-community ones. This provides a preliminary guarantee that our embeddings are ``meaningful" with respect to distance metrics. 
% It is worth mentioning that although the model size is not included explicitly in our data during training, as larger models are usually more capable in general, it can act as an indicator of models' overall abilities which is expressed in the correctness data. 

% \begin{table}[h!]
% \centering
% \begin{tabular}{ c|c|c }
%  % \hline
%  Community & Averaged Intra-Community L2 Distance & Averaged Inter-Community L2 Distance\\
%  \hline
%  7B & 9.600 & 9.624 \\
%  \hline
%  13B & 8.917 & 9.327 \\
%  \hline
%  70B & 9.219 & 9.474 \\
%  \hline
%  Coding & 9.460 & 9.591 \\
%  \hline
%  Bio/Med & 8.937 & 9.314 \\
%  \hline
%  Physics & 8.509 & 9.319 \\
%  \hline
% \end{tabular}
% \caption{The averaged intra-community L2 distance is closer for all 6 communities selected.}
% \label{table:sanity_check_experiment}
% \end{table}

\subsection{embeddings Capture Intrinsic Characteristics of Benchmarks}

Next, as indicated from \autoref{fig:regression_example}, as a set of model embeddings is produced from a fixed training set, the embeddings seem to capture information of some benchmarks better and overlook information in some benchmarks. Hence, we design a set of ablation experiments to further understand the contribution of each benchmarks in the training data.  Specifically, extending the experiment setup from \autoref{sec:bench_pred},we have a question embedding tensor $X$ of shape $(\operatorname{num\_questions}, \operatorname{embedding\_dimension})$ and a label tensor $Y$ where $Y_{ij}$ is the binary label of whether model $i$ correctly answers question $j$, with questions from a set of benchmarks $S=\{B_1, B_2 \cdots B_n\}$, to measure the effect of incorporating/removing an ``contributor" benchmark $B_i$ on predicting correctness of a ``testee" benchmark $B_j$, we:
\begin{enumerate}
    \item Construct two  sets of benchmarks $S_{added}=S \setminus B_j$ and $S_{removed} = S \setminus (B_i \cup B_j) $ and produce two new sets of question embedding and label tensor, $X_{added}$, $X_{removed}$, $Y_{added}$, and $Y_{removed}$, so that only questions containing in $S_{added}$ and $S_{removed}$ are kept respectively.
    \item Train a Matrix Factorization embedder separately on $(X_{added}, Y_{added})$ and $(X_{removed}, Y_{removed})$ to get two sets of model embeddings $E_{added}$ and $E_{removed}$, and respectively perform zero-shot benchmark prediction on $B_j$ with 100 random splits of models as in \autoref{sec:bench_pred}. Aggregate the total test mean squared error (MSE) $e_{removed}$ and $e_{added}$.
    \item Take the difference between the two error to compute a contribution score $C_{ij} = e_{removed} - e_{added}$ which quantifies the improvement on predicting model accuracy on $B_j$ when $B_i$ is added in training. 
\end{enumerate}

 % In other words, each benchmark ``contribute" to a different extent to the formation of the embedding. How can we quantify this contribution? 

% Extending the experiment setup from \ref{sec:bench_pred}, as we fix a test benchmark, we can actually further understand the ``contribution'' of each of the training benchmarks on predicting accuracies on the test benchmark. More specifically, 

% In other words, if the selected benchmark has similar traits as the test benchmark (for example both evaluating mathematics ability), then as we include the selected benchmark into training, the embedding would contain more information useful to boost performances in predicting the test benchmark's accuracy. 

Essentially, we hypothesize that the addition/removal of every training benchmark would be reflected through the change in model embeddings which induces a performance difference in benchmark accuracy prediction. With this setup, we produce a $n\times n$ ($n$ is the total number of benchmarks) contribution matrix $C$ where the where the \(ij\)-th entry can be extracted exactly as $C_{ij}$ from the steps above\footnote{The diagonal entries of this matrix are set default to 0 as it is meaningless to train and test on the same benchmark.}. To aggregate the overall effect of one specific benchmark $B_i$ as the contributor benchmark, we compute a row sum of the contribution matrix
$\sum_{j} C_{ij}$. This can be interpreted as the total improvements of adding benchmark $B_i$ on predicting model correctness of the rest of the benchmarks. Correspondingly, the column sum $\sum_{i} C_{ij}$ represents the total amount of improvement of predicting correctness of benchmark $B_j$ when each of the rest of the benchmarks is added.

Some noticeable phenomenon emerged from this set of experiments\footnote{We omit results of SocialQA in probing experiments as most of our models perform similarly poorly, resulting in all model accuracies crowding in the low accuracy region and becoming indistinguishable.}: 
\begin{enumerate}
    \item We find that incorporating MMLU into training the embeddings significantly help predicting accuracies on other benchmarks. This result matches the comprehensive nature of MMLU as it contains questions covering various topics.
    \begin{figure}[h]
        \centering
        \includegraphics[width=0.45\linewidth]{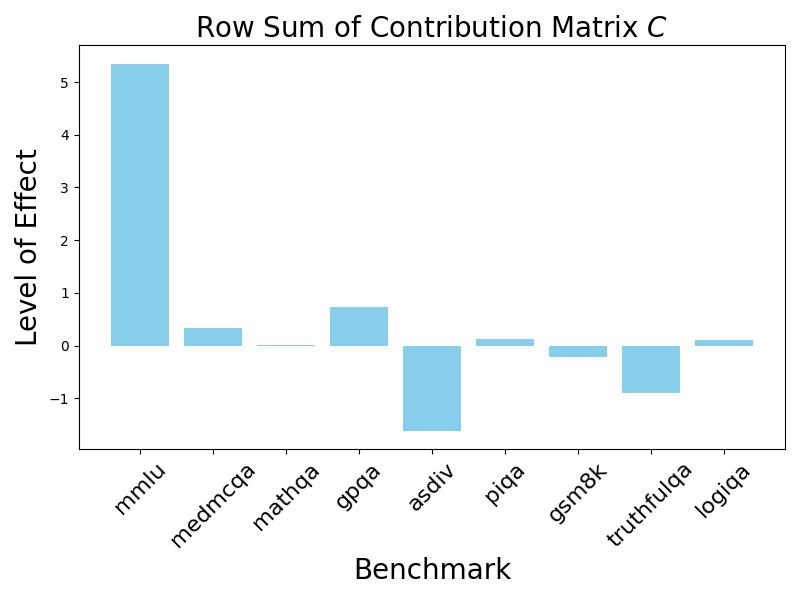}
        \includegraphics[width=0.45\linewidth]{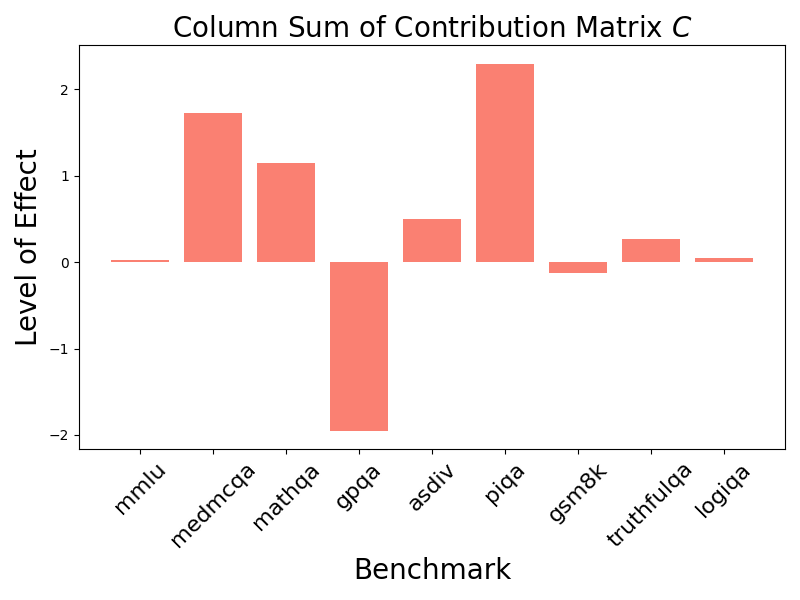}
        \caption{\textbf{Left:} Effect of removing benchmarks on testing all other benchmarks. Higher value suggests that the addition of that benchmark into the training of model embedding enhances predicting model correctness on the rest of the benchmarks. \textbf{Right:} Effect of removing each of the rest of the all other benchmarks on the benchmark being tested. Higher value suggests that predicting models' correctness in answering questions from the tested benchmark becomes easier with each of the rest of the benchmarks added overall.}
        \label{fig:heatmap_row_sum}
    \end{figure}

    \item We find that incorporating other benchmarks into training set would harm the embedding's predictive power on GPQA. This suggests that additional information that the embeddings capture from incorporating new benchmarks into training set is unrelated or negatively related to model performance on GPQA. In fact, GPQA is an extremely difficult benchmark for current LLMs, so this finding aligns with our expectation as model's ability in answering simpler questions clearly do not transfer to answering harder ones.

    \item Additionally, we identify subsets of benchmarks that mutually improve each other's accuracies when incorporated into training. For instance, there is a total MSE improvement of 0.271 when GSM8k is incorporated to predicting MathQA, 0.190 when MathQA is incorporated to predict ASDiv, and 0.103 when ASDiv is incorporated to predict MathQA. As all three benchmarks are math-related, we can deduce that in this specific subset of experiments, the level of math knowledge of our selected models are indeed reflected through the change in model embeddings.
\end{enumerate}
\section{Limitation}
Despite the promising results, our work has several limitations. First, our dataset, though effective in demonstrating the potential of our embeddings with a limited number of samples, is relatively small. With data from only 112 models, the embeddings we extract are moderately sparse which limits deeper exploration of relationships between them. Second, while our embeddings effectively support various downstream tasks, they rely on a fixed pool of models. Introducing new models requires retraining the Matrix Factorization algorithm. Lastly, our study is restricted to correctness-based datasets, leaving other potentially valuable data types, such as text embeddings of model outputs, unexplored. To address these limitations, we have open-sourced our datasets and codebase for further research and experimentation.
\section{Conclusion}

We showcase the possibility of learning an unified, compact representation of LLMs via Matrix Factorization. Through extensive empirical evaluation, our method displays solid performance on correctness forecasting, model routing, and benchmark accuracy prediction, while significantly reducing the need of retraining and avoiding repetitive evaluations. Furthermore, we conduct various probing experiment to understand the information contained in the model embeddings. The results show that our embeddings capture not only key characteristics of the models, but also properties of the data used to train the embedder. %By going into the reverse direction and using the model embedding to disclose properties of the benchmarks that are aligned with the ground truth nature of these benchmarks, we demonstrate that our methodology effectively help compresses useful information into the model embedding.

% Despite the promising results, our work has several limitations. First, our dataset, though effective in demonstrating the potential of our embeddings with a limited number of samples, is relatively small. With data from only 112 models, the embeddings we extract are moderately sparse which limits deeper exploration of relationships between them. Second, while our embeddings effectively support various downstream tasks, they rely on a fixed pool of models. Introducing new models requires retraining the Matrix Factorization algorithm. Lastly, our study is restricted to correctness-based datasets, leaving other potentially valuable data types, such as text embeddings of model outputs, unexplored. To address these limitations, we have open-sourced our datasets and codebase for further research and experimentation.

\newpage
\bibliography{iclr2025_conference}

\begin{thebibliography}{39}
\providecommand{\natexlab}[1]{#1}
\providecommand{\url}[1]{\texttt{#1}}
\expandafter\ifx\csname urlstyle\endcsname\relax
  \providecommand{\doi}[1]{doi: #1}\else
  \providecommand{\doi}{doi: \begingroup \urlstyle{rm}\Url}\fi

\bibitem[Amini et~al.(2019)Amini, Gabriel, Lin, Koncel-Kedziorski, Choi, and Hajishirzi]{amini-etal-2019-mathqa}
Aida Amini, Saadia Gabriel, Shanchuan Lin, Rik Koncel-Kedziorski, Yejin Choi, and Hannaneh Hajishirzi.
\newblock {M}ath{QA}: Towards interpretable math word problem solving with operation-based formalisms.
\newblock In \emph{Proceedings of the 2019 Conference of the North {A}merican Chapter of the Association for Computational Linguistics: Human Language Technologies, Volume 1 (Long and Short Papers)}, pp.\  2357--2367, Minneapolis, Minnesota, June 2019. Association for Computational Linguistics.
\newblock \doi{10.18653/v1/N19-1245}.
\newblock URL \url{https://aclanthology.org/N19-1245}.

\bibitem[Bisk et~al.(2019)Bisk, Zellers, Bras, Gao, and Choi]{bisk2019piqareasoningphysicalcommonsense}
Yonatan Bisk, Rowan Zellers, Ronan~Le Bras, Jianfeng Gao, and Yejin Choi.
\newblock Piqa: Reasoning about physical commonsense in natural language, 2019.
\newblock URL \url{https://arxiv.org/abs/1911.11641}.

\bibitem[Cheng et~al.(2024)Cheng, Huang, and Wei]{cheng2024adapting}
Daixuan Cheng, Shaohan Huang, and Furu Wei.
\newblock Adapting large language models via reading comprehension.
\newblock In \emph{The Twelfth International Conference on Learning Representations}, 2024.
\newblock URL \url{https://openreview.net/forum?id=y886UXPEZ0}.

\bibitem[Chiang et~al.(2024)Chiang, Zheng, Sheng, Angelopoulos, Li, Li, Zhang, Zhu, Jordan, Gonzalez, and Stoica]{chiang2024chatbotarenaopenplatform}
Wei-Lin Chiang, Lianmin Zheng, Ying Sheng, Anastasios~Nikolas Angelopoulos, Tianle Li, Dacheng Li, Hao Zhang, Banghua Zhu, Michael Jordan, Joseph~E. Gonzalez, and Ion Stoica.
\newblock Chatbot arena: An open platform for evaluating llms by human preference, 2024.
\newblock URL \url{https://arxiv.org/abs/2403.04132}.

\bibitem[Cobbe et~al.(2021)Cobbe, Kosaraju, Bavarian, Chen, Jun, Kaiser, Plappert, Tworek, Hilton, Nakano, Hesse, and Schulman]{cobbe2021gsm8k}
Karl Cobbe, Vineet Kosaraju, Mohammad Bavarian, Mark Chen, Heewoo Jun, Lukasz Kaiser, Matthias Plappert, Jerry Tworek, Jacob Hilton, Reiichiro Nakano, Christopher Hesse, and John Schulman.
\newblock Training verifiers to solve math word problems.
\newblock \emph{arXiv preprint arXiv:2110.14168}, 2021.

\bibitem[Dubois et~al.(2024)Dubois, Galambosi, Liang, and Hashimoto]{dubois2024length}
Yann Dubois, Bal{\'a}zs Galambosi, Percy Liang, and Tatsunori~B Hashimoto.
\newblock Length-controlled alpacaeval: A simple way to debias automatic evaluators.
\newblock \emph{arXiv preprint arXiv:2404.04475}, 2024.

\bibitem[Fix(1985)]{fix1985discriminatory}
Evelyn Fix.
\newblock \emph{Discriminatory analysis: nonparametric discrimination, consistency properties}, volume~1.
\newblock USAF school of Aviation Medicine, 1985.

\bibitem[Frohberg \& Binder(2022)Frohberg and Binder]{frohberg2022crassnoveldataset}
Jörg Frohberg and Frank Binder.
\newblock Crass: A novel data set and benchmark to test counterfactual reasoning of large language models, 2022.
\newblock URL \url{https://arxiv.org/abs/2112.11941}.

\bibitem[Gao et~al.(2023)Gao, Tow, Abbasi, Biderman, Black, DiPofi, Foster, Golding, Hsu, Le~Noac'h, Li, McDonell, Muennighoff, Ociepa, Phang, Reynolds, Schoelkopf, Skowron, Sutawika, Tang, Thite, Wang, Wang, and Zou]{eval-harness}
Leo Gao, Jonathan Tow, Baber Abbasi, Stella Biderman, Sid Black, Anthony DiPofi, Charles Foster, Laurence Golding, Jeffrey Hsu, Alain Le~Noac'h, Haonan Li, Kyle McDonell, Niklas Muennighoff, Chris Ociepa, Jason Phang, Laria Reynolds, Hailey Schoelkopf, Aviya Skowron, Lintang Sutawika, Eric Tang, Anish Thite, Ben Wang, Kevin Wang, and Andy Zou.
\newblock A framework for few-shot language model evaluation, 12 2023.
\newblock URL \url{https://zenodo.org/records/10256836}.

\bibitem[Hao et~al.(2022)Hao, Song, Dong, Huang, Chi, Wang, Ma, and Wei]{hao2022languagemodelsgeneralpurposeinterfaces}
Yaru Hao, Haoyu Song, Li~Dong, Shaohan Huang, Zewen Chi, Wenhui Wang, Shuming Ma, and Furu Wei.
\newblock Language models are general-purpose interfaces, 2022.
\newblock URL \url{https://arxiv.org/abs/2206.06336}.

\bibitem[Hartvigsen et~al.(2022)Hartvigsen, Gabriel, Palangi, Sap, Ray, and Kamar]{hartvigsen2022toxigenlargescalemachinegenerateddataset}
Thomas Hartvigsen, Saadia Gabriel, Hamid Palangi, Maarten Sap, Dipankar Ray, and Ece Kamar.
\newblock Toxigen: A large-scale machine-generated dataset for adversarial and implicit hate speech detection, 2022.
\newblock URL \url{https://arxiv.org/abs/2203.09509}.

\bibitem[He et~al.(2022)He, Chen, Xie, Li, Doll\'ar, and Girshick]{He_2022_CVPR_MAE}
Kaiming He, Xinlei Chen, Saining Xie, Yanghao Li, Piotr Doll\'ar, and Ross Girshick.
\newblock Masked autoencoders are scalable vision learners.
\newblock In \emph{Proceedings of the IEEE/CVF Conference on Computer Vision and Pattern Recognition (CVPR)}, pp.\  16000--16009, June 2022.

\bibitem[Hendrycks et~al.(2021)Hendrycks, Burns, Basart, Zou, Mazeika, Song, and Steinhardt]{hendryckstest2021MMLU}
Dan Hendrycks, Collin Burns, Steven Basart, Andy Zou, Mantas Mazeika, Dawn Song, and Jacob Steinhardt.
\newblock Measuring massive multitask language understanding.
\newblock \emph{Proceedings of the International Conference on Learning Representations (ICLR)}, 2021.

\bibitem[Hu et~al.(2024)Hu, Bieker, Li, Jiang, Keigwin, Ranganath, Keutzer, and Upadhyay]{hu2024routerbenchbenchmarkmultillmrouting}
Qitian~Jason Hu, Jacob Bieker, Xiuyu Li, Nan Jiang, Benjamin Keigwin, Gaurav Ranganath, Kurt Keutzer, and Shriyash~Kaustubh Upadhyay.
\newblock Routerbench: A benchmark for multi-llm routing system, 2024.
\newblock URL \url{https://arxiv.org/abs/2403.12031}.

\bibitem[Li et~al.(2024)Li, Chiang, Frick, Dunlap, Wu, Zhu, Gonzalez, and Stoica]{li2024crowdsourced}
Tianle Li, Wei-Lin Chiang, Evan Frick, Lisa Dunlap, Tianhao Wu, Banghua Zhu, Joseph~E Gonzalez, and Ion Stoica.
\newblock From crowdsourced data to high-quality benchmarks: Arena-hard and benchbuilder pipeline.
\newblock \emph{arXiv preprint arXiv:2406.11939}, 2024.

\bibitem[Lin et~al.(2022)Lin, Hilton, and Evans]{lin2022truthfulqameasuringmodelsmimic}
Stephanie Lin, Jacob Hilton, and Owain Evans.
\newblock Truthfulqa: Measuring how models mimic human falsehoods, 2022.
\newblock URL \url{https://arxiv.org/abs/2109.07958}.

\bibitem[Liu et~al.(2020)Liu, Cui, Liu, Huang, Wang, and Zhang]{liu2020logiqachallengedatasetmachine}
Jian Liu, Leyang Cui, Hanmeng Liu, Dandan Huang, Yile Wang, and Yue Zhang.
\newblock Logiqa: A challenge dataset for machine reading comprehension with logical reasoning, 2020.
\newblock URL \url{https://arxiv.org/abs/2007.08124}.

\bibitem[Lu et~al.(2023)Lu, Yuan, Lin, Lin, Yuan, Zhou, and Zhou]{lu2023routingexpertefficientrewardguided}
Keming Lu, Hongyi Yuan, Runji Lin, Junyang Lin, Zheng Yuan, Chang Zhou, and Jingren Zhou.
\newblock Routing to the expert: Efficient reward-guided ensemble of large language models, 2023.
\newblock URL \url{https://arxiv.org/abs/2311.08692}.

\bibitem[Miao et~al.(2020)Miao, Liang, and Su]{asdiv}
Shen-yun Miao, Chao-Chun Liang, and Keh-Yih Su.
\newblock A diverse corpus for evaluating and developing english math word problem solvers.
\newblock In \emph{Proceedings of the 58th Annual Meeting of the Association for Computational Linguistics}, pp.\  975--984, 2020.

\bibitem[Mikolov et~al.(2013{\natexlab{a}})Mikolov, Chen, Corrado, and Dean]{mikolov2013efficientestimationwordrepresentations}
Tomas Mikolov, Kai Chen, Greg Corrado, and Jeffrey Dean.
\newblock Efficient estimation of word representations in vector space, 2013{\natexlab{a}}.
\newblock URL \url{https://arxiv.org/abs/1301.3781}.

\bibitem[Mikolov et~al.(2013{\natexlab{b}})Mikolov, Yih, and Zweig]{mikolov-etal-2013-linguistic}
Tomas Mikolov, Wen-tau Yih, and Geoffrey Zweig.
\newblock Linguistic regularities in continuous space word representations.
\newblock In Lucy Vanderwende, Hal Daum{\'e}~III, and Katrin Kirchhoff (eds.), \emph{Proceedings of the 2013 Conference of the North {A}merican Chapter of the Association for Computational Linguistics: Human Language Technologies}, pp.\  746--751, Atlanta, Georgia, June 2013{\natexlab{b}}. Association for Computational Linguistics.
\newblock URL \url{https://aclanthology.org/N13-1090}.

\bibitem[Noroozi \& Favaro(2017)Noroozi and Favaro]{noroozi2017unsupervisedlearningvisualrepresentations}
Mehdi Noroozi and Paolo Favaro.
\newblock Unsupervised learning of visual representations by solving jigsaw puzzles, 2017.
\newblock URL \url{https://arxiv.org/abs/1603.09246}.

\bibitem[Ong et~al.(2024)Ong, Almahairi, Wu, Chiang, Wu, Gonzalez, Kadous, and Stoica]{ong2024routellm}
Isaac Ong, Amjad Almahairi, Vincent Wu, Wei-Lin Chiang, Tianhao Wu, Joseph~E Gonzalez, M~Waleed Kadous, and Ion Stoica.
\newblock Routellm: Learning to route llms with preference data.
\newblock \emph{arXiv preprint arXiv:2406.18665}, 2024.

\bibitem[Pal et~al.(2022)Pal, Umapathi, and Sankarasubbu]{medmcqa}
Ankit Pal, Logesh~Kumar Umapathi, and Malaikannan Sankarasubbu.
\newblock Medmcqa: A large-scale multi-subject multi-choice dataset for medical domain question answering.
\newblock In Gerardo Flores, George~H Chen, Tom Pollard, Joyce~C Ho, and Tristan Naumann (eds.), \emph{Proceedings of the Conference on Health, Inference, and Learning}, volume 174 of \emph{Proceedings of Machine Learning Research}, pp.\  248--260. PMLR, 07--08 Apr 2022.
\newblock URL \url{https://proceedings.mlr.press/v174/pal22a.html}.

\bibitem[Pennington et~al.(2014)Pennington, Socher, and Manning]{pennington-etal-2014-glove}
Jeffrey Pennington, Richard Socher, and Christopher Manning.
\newblock {G}lo{V}e: Global vectors for word representation.
\newblock In Alessandro Moschitti, Bo~Pang, and Walter Daelemans (eds.), \emph{Proceedings of the 2014 Conference on Empirical Methods in Natural Language Processing ({EMNLP})}, pp.\  1532--1543, Doha, Qatar, October 2014. Association for Computational Linguistics.
\newblock \doi{10.3115/v1/D14-1162}.
\newblock URL \url{https://aclanthology.org/D14-1162}.

\bibitem[Reimers \& Gurevych(2019)Reimers and Gurevych]{reimers-2019-sentence-bert}
Nils Reimers and Iryna Gurevych.
\newblock Sentence-bert: Sentence embeddings using siamese bert-networks.
\newblock In \emph{Proceedings of the 2019 Conference on Empirical Methods in Natural Language Processing}. Association for Computational Linguistics, 11 2019.
\newblock URL \url{https://arxiv.org/abs/1908.10084}.

\bibitem[Rein et~al.(2023)Rein, Hou, Stickland, Petty, Pang, Dirani, Michael, and Bowman]{rein2023gpqagraduatelevelgoogleproofqa}
David Rein, Betty~Li Hou, Asa~Cooper Stickland, Jackson Petty, Richard~Yuanzhe Pang, Julien Dirani, Julian Michael, and Samuel~R. Bowman.
\newblock Gpqa: A graduate-level google-proof q\&a benchmark, 2023.
\newblock URL \url{https://arxiv.org/abs/2311.12022}.

\bibitem[Ronneberger et~al.(2015)Ronneberger, Fischer, and Brox]{ronneberger2015unetconvolutionalnetworksbiomedical}
Olaf Ronneberger, Philipp Fischer, and Thomas Brox.
\newblock U-net: Convolutional networks for biomedical image segmentation, 2015.
\newblock URL \url{https://arxiv.org/abs/1505.04597}.

\bibitem[Sap et~al.(2019)Sap, Rashkin, Chen, LeBras, and Choi]{sap2019socialiqacommonsensereasoningsocial}
Maarten Sap, Hannah Rashkin, Derek Chen, Ronan LeBras, and Yejin Choi.
\newblock Socialiqa: Commonsense reasoning about social interactions, 2019.
\newblock URL \url{https://arxiv.org/abs/1904.09728}.

\bibitem[Shao et~al.(2024)Shao, Wang, Zhu, Xu, Song, Bi, Zhang, Zhang, Li, Wu, and Guo]{shao2024deepseekmathpushinglimitsmathematical}
Zhihong Shao, Peiyi Wang, Qihao Zhu, Runxin Xu, Junxiao Song, Xiao Bi, Haowei Zhang, Mingchuan Zhang, Y.~K. Li, Y.~Wu, and Daya Guo.
\newblock Deepseekmath: Pushing the limits of mathematical reasoning in open language models, 2024.
\newblock URL \url{https://arxiv.org/abs/2402.03300}.

\bibitem[Shnitzer et~al.(2023)Shnitzer, Ou, Silva, Soule, Sun, Solomon, Thompson, and Yurochkin]{shnitzer2023largelanguagemodelrouting}
Tal Shnitzer, Anthony Ou, Mírian Silva, Kate Soule, Yuekai Sun, Justin Solomon, Neil Thompson, and Mikhail Yurochkin.
\newblock Large language model routing with benchmark datasets, 2023.
\newblock URL \url{https://arxiv.org/abs/2309.15789}.

\bibitem[Srinivasan et~al.(2023)Srinivasan, Dong, Zhu, Yu, Mosk-Aoyama, Keutzer, Jiao, and Zhang]{srinivasan2023nexusraven}
Venkat~Krishna Srinivasan, Zhen Dong, Banghua Zhu, Brian Yu, Damon Mosk-Aoyama, Kurt Keutzer, Jiantao Jiao, and Jian Zhang.
\newblock Nexusraven: a commercially-permissive language model for function calling.
\newblock In \emph{NeurIPS 2023 Foundation Models for Decision Making Workshop}, 2023.

\bibitem[Vaswani et~al.(2023)Vaswani, Shazeer, Parmar, Uszkoreit, Jones, Gomez, Kaiser, and Polosukhin]{vaswani2023attentionneed}
Ashish Vaswani, Noam Shazeer, Niki Parmar, Jakob Uszkoreit, Llion Jones, Aidan~N. Gomez, Lukasz Kaiser, and Illia Polosukhin.
\newblock Attention is all you need, 2023.
\newblock URL \url{https://arxiv.org/abs/1706.03762}.

\bibitem[Vondrick et~al.(2018)Vondrick, Shrivastava, Fathi, Guadarrama, and Murphy]{vondrick2018trackingemergescolorizingvideos}
Carl Vondrick, Abhinav Shrivastava, Alireza Fathi, Sergio Guadarrama, and Kevin Murphy.
\newblock Tracking emerges by colorizing videos, 2018.
\newblock URL \url{https://arxiv.org/abs/1806.09594}.

\bibitem[Wang et~al.(2019)Wang, Singh, Michael, Hill, Levy, and Bowman]{wang2019gluemultitaskbenchmarkanalysis}
Alex Wang, Amanpreet Singh, Julian Michael, Felix Hill, Omer Levy, and Samuel~R. Bowman.
\newblock Glue: A multi-task benchmark and analysis platform for natural language understanding, 2019.
\newblock URL \url{https://arxiv.org/abs/1804.07461}.

\bibitem[Yan et~al.(2024)Yan, Mao, Ji, Zhang, Patil, Stoica, and Gonzalez]{berkeley-function-calling-leaderboard}
Fanjia Yan, Huanzhi Mao, Charlie Cheng-Jie Ji, Tianjun Zhang, Shishir~G. Patil, Ion Stoica, and Joseph~E. Gonzalez.
\newblock Berkeley function calling leaderboard.
\newblock 2024.

\bibitem[Yu et~al.(2023)Yu, Jiang, Shi, Yu, Liu, Zhang, Kwok, Li, Weller, and Liu]{yu2023metamath}
Longhui Yu, Weisen Jiang, Han Shi, Jincheng Yu, Zhengying Liu, Yu~Zhang, James~T Kwok, Zhenguo Li, Adrian Weller, and Weiyang Liu.
\newblock Metamath: Bootstrap your own mathematical questions for large language models.
\newblock \emph{arXiv preprint arXiv:2309.12284}, 2023.

\bibitem[Zheng et~al.(2023)Zheng, Chiang, Sheng, Zhuang, Wu, Zhuang, Lin, Li, Li, Xing, Zhang, Gonzalez, and Stoica]{zheng2023judging}
Lianmin Zheng, Wei-Lin Chiang, Ying Sheng, Siyuan Zhuang, Zhanghao Wu, Yonghao Zhuang, Zi~Lin, Zhuohan Li, Dacheng Li, Eric.~P Xing, Hao Zhang, Joseph~E. Gonzalez, and Ion Stoica.
\newblock Judging llm-as-a-judge with mt-bench and chatbot arena, 2023.

\bibitem[Zhu et~al.(2024)Zhu, Frick, Wu, Zhu, Ganesan, Chiang, Zhang, and Jiao]{zhu2024starling}
Banghua Zhu, Evan Frick, Tianhao Wu, Hanlin Zhu, Karthik Ganesan, Wei-Lin Chiang, Jian Zhang, and Jiantao Jiao.
\newblock Starling-7b: Improving helpfulness and harmlessness with rlaif.
\newblock In \emph{First Conference on Language Modeling}, 2024.

\end{thebibliography}
\bibliographystyle{iclr2025_conference}

\appendix

\section{Appendix}

\subsection{Model List}

Here is an exhaustive list of models that we extract our dataset from:
\begin{table}[h!]
\centering
\begin{tabular}{|c|c|}
\hline
meta-llama/LlamaGuard-7b & meta-llama/Llama-2-13b-chat-hf \\
01-ai/Yi-34B-Chat & meta-llama/Llama-2-70b-chat-hf \\
WizardLM/WizardLM-70B-V1.0 & allenai/tulu-2-dpo-70b \\
lmsys/vicuna-13b-v1.5 & lmsys/vicuna-33b-v1.3 \\
Qwen/Qwen-14B-Chat & upstage/SOLAR-10.7B-Instruct-v1.0 \\
openchat/openchat-3.5-0106 & openchat/openchat-3.5 \\
berkeley-nest/Starling-LM-7B-alpha & HuggingFaceH4/zephyr-7b-beta \\
TheBloke/tulu-30B-fp16 & mistralai/Mistral-7B-Instruct-v0.1 \\
tiiuae/falcon-40b-instruct & lmsys/vicuna-13b-v1.5-16k \\
codellama/CodeLlama-34b-Instruct-hf & TheBloke/WizardLM-13B-V1.2-GGUF \\
lmsys/vicuna-7b-v1.5 & NousResearch/Nous-Hermes-13b \\
project-baize/baize-v2-13b & lmsys/vicuna-7b-v1.5-16k \\
mosaicml/mpt-30b-instruct & meta-llama/Llama-2-7b-chat-hf \\
TheBloke/koala-13B-HF & nomic-ai/gpt4all-13b-snoozy \\
h2oai/h2ogpt-gm-oasst1-en-2048-open-llama-13b & mosaicml/mpt-7b-chat \\
databricks/dolly-v2-12b & stabilityai/stablelm-tuned-alpha-7b \\
OpenAssistant/oasst-sft-4-pythia-12b-epoch-3.5 & deepseek-ai/deepseek-llm-67b-chat \\
NousResearch/Nous-Hermes-2-Yi-34B & CausalLM/34b-beta \\
SUSTech/SUS-Chat-34B & SUSTech/SUS-Chat-72B \\
Qwen/Qwen-72B & Intel/neural-chat-7b-v3-3 \\
ibivibiv/alpaca-dragon-72b-v1 & JaeyeonKang/CCK-Asura-v1 \\
ConvexAI/Luminex-34B-v0.2 & ConvexAI/Luminex-34B-v0.1 \\
CorticalStack/pastiche-crown-clown-7b-dare-dpo & eren23/ogno-monarch-jaskier-merge-7b-OH-PREF-DPO \\
bardsai/jaskier-7b-dpo-v5.6 & FelixChao/Scorpio-7B \\
dfurman/HermesBagel-34B-v0.1 & kevin009/llamaRAGdrama \\
sail/Sailor-7B & AiMavenAi/Prometheus-1.3 \\
Q-bert/Optimus-7B & cognitivecomputations/yayi2-30b-llama \\
zhengr/MixTAO-7Bx2-MoE-v8.1 & fblgit/UNA-SimpleSmaug-34b-v1beta \\
mistralai/Mixtral-8x7B-Instruct-v0.1 & microsoft/Orca-2-13b \\
EleutherAI/pythia-12b & cloudyu/Mixtral-11Bx2-MoE-19B \\
rishiraj/CatPPT-base & Deci/DeciLM-7B \\
microsoft/phi-2 & scb10x/typhoon-7b \\
01-ai/Yi-6B-200K & 01-ai/Yi-6B \\
TigerResearch/tigerbot-13b-base & augmxnt/shisa-base-7b-v1 \\
microsoft/phi-1.5 & golaxy/gowizardlm \\
bigscience/bloom-7b1 & mlabonne/AlphaMonarch-7B \\
CultriX/NeuralTrix-bf16 & shadowml/MBeagleX-7B \\
yam-peleg/Experiment26-7B & deepseek-ai/deepseek-math-7b-instruct \\
meta-math/MetaMath-Mistral-7B & kyujinpy/Sakura-SOLRCA-Math-Instruct-DPO-v1 \\
FelixChao/llama2-13b-math1.2 & Plaban81/Moe-4x7b-math-reason-code \\
MaziyarPanahi/WizardLM-Math-70B-v0.1 & abhishek/zephyr-beta-math \\
meta-math/MetaMath-Llemma-7B & EleutherAI/llemma-34b \\
EleutherAI/llemma-7b & FelixChao/vicuna-7B-physics \\
Harshvir/Llama-2-7B-physics & FelixChao/vicuna-7B-chemical \\
BioMistral/BioMistral-7B & BioMistral/BioMistral-7B-DARE \\
PharMolix/BioMedGPT-LM-7B & Biomimicry-AI/ANIMA-Nectar-v2 \\
codellama/CodeLlama-7b-hf & codellama/CodeLlama-13b-Instruct-hf \\
deepseek-ai/deepseek-coder-1.3b-base & deepseek-ai/deepseek-coder-6.7b-instruct \\
OpenBuddy/openbuddy-codellama2-34b-v11.1-bf16 & TheBloke/CodeLlama-70B-Instruct-AWQ \\
AdaptLLM/medicine-chat & AdaptLLM/medicine-LLM \\
AdaptLLM/medicine-LLM-13B & Writer/palmyra-med-20b \\
SciPhi/SciPhi-Self-RAG-Mistral-7B-32k & Neko-Institute-of-Science/metharme-7b \\
Neko-Institute-of-Science/pygmalion-7b & SciPhi/SciPhi-Mistral-7B-32k \\
shleeeee/mistral-ko-tech-science-v1 & codefuse-ai/CodeFuse-DeepSeek-33B \\
WizardLM/WizardCoder-Python-34B-V1.0 & bigcode/octocoder \\
meta-llama/Meta-Llama-3-8B & meta-llama/Meta-Llama-3-8B-Instruct \\
meta-llama/Meta-Llama-3-70B & meta-llama/Meta-Llama-3-70B-Instruct \\
meta-llama/Meta-Llama-Guard-2-8B & Qwen/Qwen1.5-32B-Chat \\
Qwen/Qwen1.5-4B-Chat & Qwen/Qwen1.5-0.5B-Chat \\
Qwen/Qwen1.5-7B-Chat & Nexusflow/Starling-LM-7B-beta \\
google/gemma-7b-it & google/gemma-2b-it \\
\hline
\end{tabular}
\caption{The comprehensive list of the 112 models used to create the correctness dataset. This dataset is created on May 2025 so models released after that time are not available.}
\end{table}

\end{document}